\pdfoutput=1

\documentclass[11pt]{article}

\usepackage{ACL2023}

\usepackage{times}
\usepackage{latexsym}

\usepackage[T1]{fontenc}

\usepackage[utf8]{inputenc}

\usepackage{microtype}

\usepackage{inconsolata}
\usepackage{stfloats}

\usepackage{ctable} 
\usepackage{subcaption}
\usepackage{physics}
\usepackage{amsmath,amsfonts,amssymb}
\usepackage{multicol}
\newsavebox{\algleft}
\newsavebox{\algright}
\usepackage{bm}
\usepackage[linesnumbered,ruled,vlined]{algorithm2e}
\SetKwComment{Comment}{/* }{ */} 
\usepackage{hyperref}       
\usepackage{url}            
\usepackage{booktabs}       
\usepackage{amsfonts}       
\usepackage{nicefrac}       
\usepackage{xcolor}         
\usepackage{multirow}
\usepackage{soul}

\graphicspath{{./figure/}}

\title{Scalable and Safe Remediation of Defective Actions\\ in Self-Learning Conversational Systems}

\author{Sarthak Ahuja, Mohammad Kachuee, Fateme Sheikholeslami, Weiqing Liu, Jaeyoung Do\\
  Amazon Alexa AI, Seattle, WA \\
  \texttt{\{sarahuja, kachum, shfateme, lweiqing, domjae\}@amazon.com} \\}

\begin{document}
\maketitle
\begin{abstract}
Off-Policy reinforcement learning has been a driving force for the state-of-the-art conversational AIs leading to more natural human-agent interactions and improving the user satisfaction for goal-oriented agents. However, in large-scale commercial settings, it is often challenging to balance between policy improvements and experience continuity on the broad spectrum of applications handled by such system. In the literature, off-policy evaluation and guard-railing on aggregate statistics has been commonly used to address this problem. In this paper, we propose a method for curating and leveraging high-precision samples sourced from historical regression incident reports to validate, safe-guard, and improve policies prior to the online deployment. We conducted extensive experiments using data from a real-world conversational system and actual regression incidents. The proposed method is currently deployed in our production system to protect customers against broken experiences and enable long-term policy improvements.
\end{abstract}
\section{Introduction}
Conversational AI systems such as Apple Siri, Amazon Alexa, Google Assistant, and Microsoft Cortana rely on multiple components for speech recognition, natural language understanding (NLU), skill routing, and generating a response to the user. A skill routing block selects the right skill/provider and NLU interpretation to serve a user's request. Skill routing is a challenging problem due to the number of skills present in a real-world conversational system. Furthermore, new skills are being introduced every day, existing skills may change behavior over time while some others getting deprecated leading to an ever changing customer-skill dynamic \citep{sarikaya2017technology,park2020scalable}.

\begin{figure}[h]
    \centering
        \includegraphics[width=\linewidth]{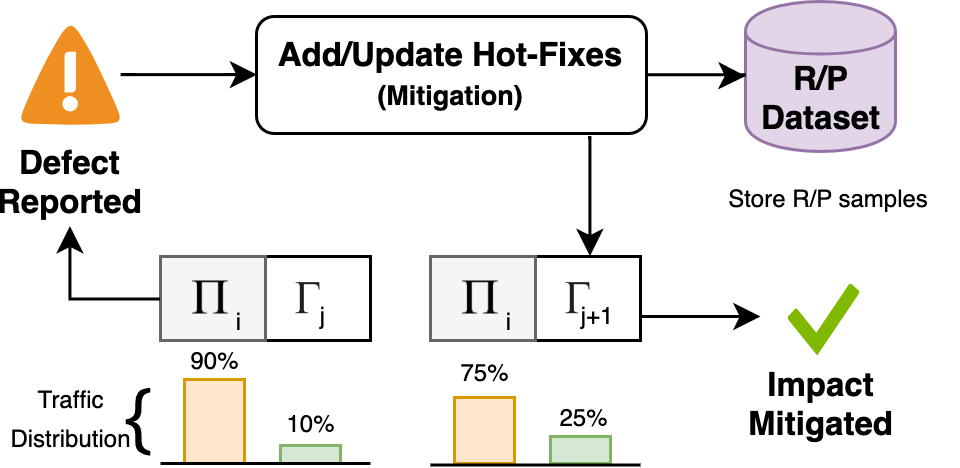}
        \caption{To immediately mitigate the business impact of a reported defect usually a high-recall hot-fix is added to the system such that the problematic traffic segment is redirected away from the RL policy ($\Pi$) towards a hand-crafted rule policy ($\Gamma$) representing this hot-fix; We propose to maintain a dataset of regression and progression samples (R/P) associated with the defect to guard-rail against future recurrence and eventually assimilate the redirected traffic back to the RL policy.}
        \label{fig:mitigation}
\end{figure}

To address such challenges, state of the art skill routing systems cast the problem as a reinforcement learning (RL) problem where the agent performs periodic off-policy updates. The RL agent continually improves or self-learns by exploring alternative decisions and learning from the logged customer interaction data \citep{kachuee2022scalable}. While the RL-based approach has many merits around scalability such as no need for expensive human annotation, it also has a tendency to cause instabilities in the agent's behavior which not only regress user retention and trust, but also manifest as revenue loss for business-critical domains \citep{kachuee2022constrained,ke2022domain}.

Any policy update inherently entails a risk of breaking certain current user experience, as each deployment despite improving the overall aggregate performance, may regress on certain sub-populations and edge cases which is not acceptable in a commercial system \citep{li2021neural}. Furthermore, the frequent and automated nature of these refreshes proportionately increases this risk for the policy to deviate from its stable state when handling edge cases. 
Pre-deployment offline evaluation and constrained optimization 
can guardrail against such regressions but are 
limited by 
predefined segmentation of data and metrics that only consider coarse sub-populations \citep{kachuee2021self,kachuee2022scalable,hoffman2014correlation,balakrishnan2018using}.

These statistical approaches to learning and evaluation further struggle to let the agent protect, learn and retain knowledge of historical regressions that are self-reported by users. Such incidents are usually characterized as belonging to a narrow traffic segment but of high importance where reward metrics are not very reliable.
Typically, to mitigate them, high-recall hot-fixes are deployed to override policy and quickly address the incident as depicted in figure \ref{fig:mitigation}. Note that these hot-fixes are often hand-crafted rules that are not reliable for guard-railing against recurrence and performing a long-term remediation \cite{karampatziakis2019lessons}. 

\begin{figure}[h]
    \centering
        \includegraphics[width=\linewidth]{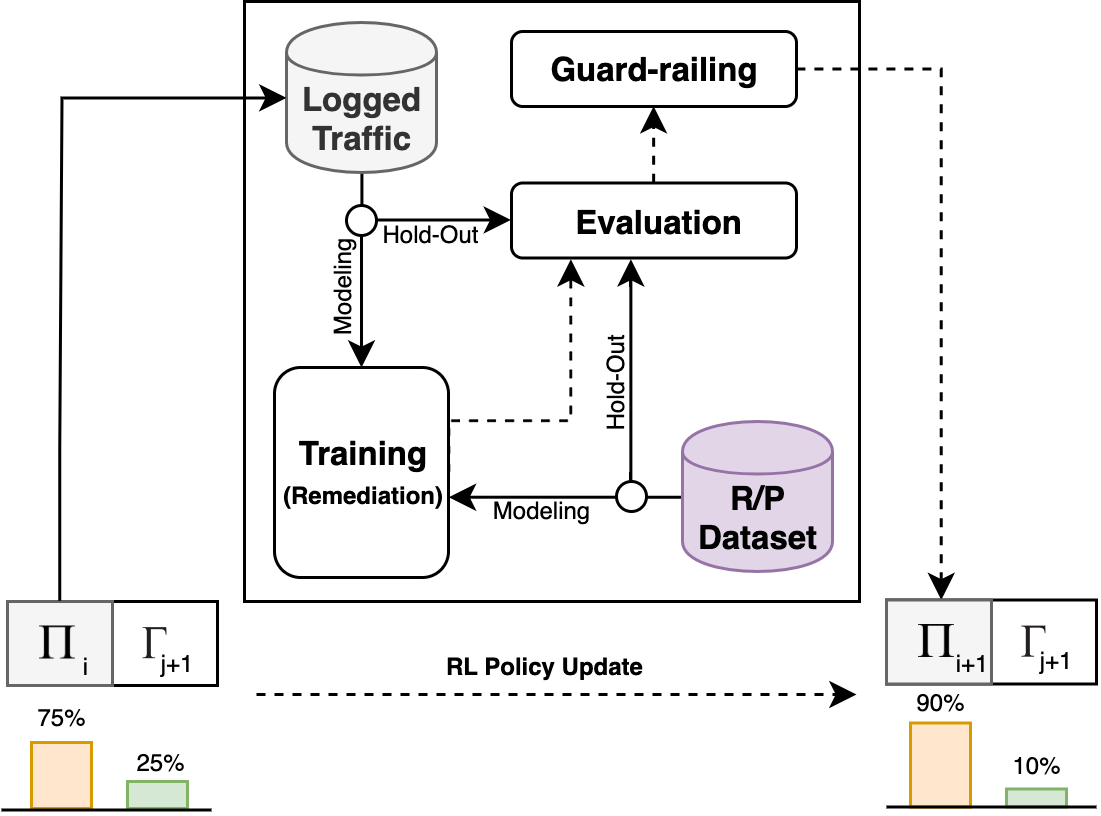}
        \caption{Post mitigation, for more permanent remediation, we leverage the R/P dataset to provide an auxiliary signal during policy updates and assimilate the instance level behavior from the samples back into policy, thereby retiring the hot-fixes over time. We promote an updated policy to production after evaluating it against test R/P data and ensuring that the resulting metrics clear a set of guard-rails that prevent recurrence of a historically reported defect.}
        \label{fig:overview}
        \vspace{-0.1in}
\end{figure}

In this paper we posit that for business-critical user-reported defects it is crucial to consider individual cases so as to learn and gate on the instance-level behavior directly. In other words, we propose complementing the current learning and evaluation mechanisms operating on aggregate metrics with high-precision instance-level analysis.
Herein, we outline a novel architecture that extends RL-based skill-routing to use a set of curated high-value user-reported defective samples, for guard-railing against re-occurrence and performing long-term remediation to re-onboard those cases to the policy; thereby retiring the hot-fixing rules introduced during the short-term mitigation. A high-level overview of the proposed system is presented in figure \ref{fig:overview}. 

To evaluate the suggested framework, we conducted extensive online and offline experiments using data from a real-world conversational agent.
We observe that the proposed approach leads to a high assimilation ($>70\%$) of the defective traffic back to RL policy i.e. long-term remediation and eventual retirement of the hot-fixes. Further, the deviation percentage in decision replication rate and the expected reward in both offline and online settings indicate that the proposed approach has no statistically significant side-effect on the remaining traffic segments.



\section{Proposed Method}

\subsection{Problem Formulation}
\label{sec:problem}
We consider the general formulation for an RL agent characterized by $\Pi_{\theta}(a|X)$ where $\theta$ are trainable parameters to specify the action selection distribution for each action $a \in \{1\dots T\}$  conditioned on the current state/context, $X$.
Here, after taking an action, the agent observes a reward denoted by $r$.
The task for the agent is to learn from the experiences collected from the current policy, $\Pi_0(a|X)$, interactions in an off-policy setting, to train a new policy parameterized by $\theta$, $\Pi_{\theta}(a|X)$.

Off-policy updates are not always stable and occasionally lead to unsatisfactory 
decisions \citep{swaminathan2016off,joachims2018deep,lopez2021learning}. These incidents are reported 
in the form of a handful of samples reproducing the defective action called \textit{regression} samples. Alongside the regression samples, typically, the report is further supplemented with complementary and contrasting samples by the user that convey the desired agent behavior. Such samples are referred to as \textit{progression} samples here. 
Collectively we denote the dataset of all such reported regression and progression (R/P) samples across all incidents as $\mathbb{D}_{RP}$. These high value samples are carefully stored with additional meta-data 
and used in evaluating against their recurrence of these incidents (section \ref{sec:evaluation}) as well as for their long-term remediation by getting assimilated into the policy (section \ref{sec:remediation}). The meta-data may contain information such as unique sample identifiers, description of the issue, type of the sample (i.e. regression or progression), severity of the corresponding incident, date which the sample was reported, and the current life-cycle status of the sample (i.e. deprecated or active).


Remediation involves providing supervision signals for policy updates which is a non-trivial and time-consuming process. Meanwhile, to immediately mitigate business impact from an incident, hot-fixing is usually employed by introducing hand-crafted rules on the problematic segment. The set of hand-crafted rules from all incidents reported in a time period, define an eligibility criteria, $G(\Pi_\theta,X)$ that decides based on the input sample $X$ and the associated policy $\Pi_\theta$, if an input sample is eligible for the RL policy or should be handled by the hand-crafted rules. We use the notation $G(\Pi_\theta, X) \in \{0,1\}$ to represent the logic that returns one if a sample should be handled by $\Pi_\theta$, or zero if should be redirected to hot-fixes.

The set of hot-fixes can be thought of as a separate abstract policy $\Gamma(a|X)$ that runs on incoming traffic whenever the eligibility criteria $G(\Pi_\theta,X)$ is not satisfied:
\begin{equation}
      \Pi_{\theta}(a|X)=\begin{cases}
    \Gamma(a|X) & \text{$G(\Pi_\theta,X) = 0$}\\
    \Pi_{\theta}(a|X) & \text{otherwise}
  \end{cases}.
\end{equation}


\subsection{Evaluation}
\label{sec:evaluation}

The evaluation process starts by replaying the new policy $\Pi_{\theta}$ on the curated samples $(X, a, r)\in\{\mathbb{D}_{RP}\}$ to get the policy action propensities $\Pi_{\theta}(X)$. Then, we compute the most likely action under the new policy as $\widehat{a} = \arg\max(\Pi_{\theta}(X))$. 

For progression samples, we report a sample as \textit{pass} if $\widehat{a}$ is equal to the logged action $a$, otherwise it is considered as a \textit{fail} case. Alternatively, for regression samples, it would be considered as a fail if and only if the logged unsatisfactory action was repeated by the new policy. Also, to assign fail/pass certainties for each case, we compute the likelihood of each assignment as $\Pi_{\theta}(\widehat{a}|X)$ for passed progression or failed regression, and otherwise $1-\Pi_{\theta}(\widehat{a}|X)$.

Additionally, we can compute the expected eligibility of a sample given the new policy as:
\begin{align}
   Q(X):&= \mathbb{E} [G({\Pi_{\theta}},X)]  \nonumber \\ & =  \sum_{i \in 1 \dots |a|} G(\Pi_{\theta}(a_i|X)) \Pi_{\theta}(a_i|X)
\end{align}
Intuitively, $\mathbb{E} [G({\Pi_{\theta}},X)]$ measures the expected likelihood of handling sample $X$ by policy $\Pi_{\theta}$ rather than a hot-fix.

Thus  in short, we report the following evaluation metrics for each R/P sample in the evaluation stage:
\begin{enumerate}
\item \textbf{Expected Eligibility (Q)}: probability that a particular sample will be served by the RL policy given the current state of hot-fixes in place; $0 \leq P(Q) \leq 1$.
\item \textbf{Sample Status Certainty (C)}: confidence on the assigned sample status (PASS/FAIL) based on the evaluation of the policy output for that particular sample; $0 \leq P(C) \leq 1$.
\end{enumerate}

\begin{figure}[pt]
    \centering
        \includegraphics[width=0.99\linewidth]{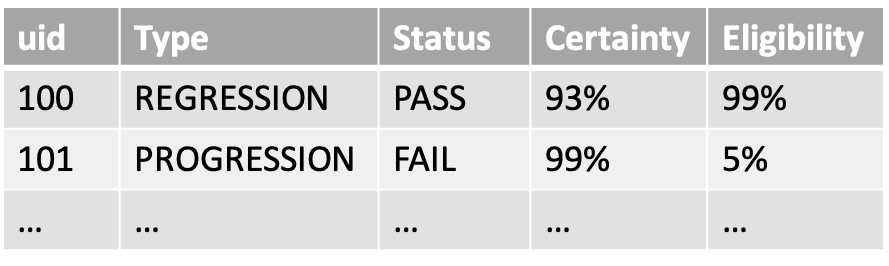}
        \caption{An example of report generated during R/P evaluation consisting of unique identifier (uid), samples type, pass/fail evaluation status, pass/fail certainty, and likelihood of handing by policy rather than hot-fixes (eligibility). In this example, the second sample failed with high certainty but since eligibility is relatively low, it would be less concerning for potential deployment.}
        \label{fig:evaluation}
\end{figure}

The last step for the evaluation is to generate a report to be used by human operators as well as automated guard-railing (next step) to understand any failures, their certainty, and likelihood of exposing such behavior to the end user. Figure~\ref{fig:evaluation} shows an example of such report.

\subsection{Guard-railing}
Hot-fixes introduced for mitigating business impact due to high-severity regression incidents are conditioned on the policy input ($X$) and the output ($\Pi_\theta(a|X)$). Thus in the event of a subsequent policy refresh, there is always a chance that the associated eligibility criteria $G(\Pi_\theta, X)$ for the associated hand-crafted rules gets out-dated and starts to redirect the problematic traffic segments to the RL model. To prevent the recurrence of the regressions, we perform pre-deployment guard-railing right after every policy update using the evaluation parameters defined in section \ref{sec:evaluation}

\begin{figure}[pt]
    \centering
        \includegraphics[width=0.9\linewidth]{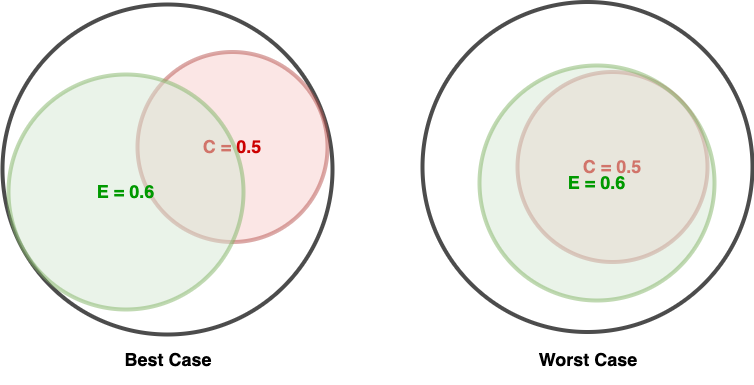}
        \caption{\textbf{left}: in the best case scenario there would be a minimal overlap between sample spaces that are eligible for the RL policy and will lead to potential defects. \textbf{right}: in the worst case scenario there would be a maximum overlap between the aforementioned sample spaces.}
        \label{fig:gating}
\end{figure}

For the sample X, assumed at index $i$ of $\mathbb{D}_{RP}$, we perform gating on their intersection probability of the experiment eligibility and sample status certainty $P(C_i \cap Q_i)$ i.e. a sample being eligible for the RL policy with a high certainty of causing a misroute. For failing cases ($C_i =$ FAIL), the best (most lenient) and worst (most strict) case scenario are depicted in figure \ref{fig:gating}. To prevent any unnecessarily blocks, we use the best case setup when comparing the minimum intersection probability against a set failure threshold $T_f$. For passing samples ($C_i =$ PASS) we simply invert the sample certainty value and keep the remaining logic as is. Algorithm \ref{alg:gating} summarizes the guard-railing logic for the failing case for a single sample. 

\begin{algorithm}[t]
\small
\caption{Guard-railing on a single failing regression/progression sample}
\label{alg:gating}

\DontPrintSemicolon
\SetKwInOut{Input}{input}
\SetKwInOut{Output}{output}

\Input{i (RP sample index),\\ $P(C=FAIL) \sim P(C) $ (failure certainty),\\ $P(Q)$ (expected eligibility), \\ $T_f$ (failure threshold for guard-railing)}

\If{$P(C_i) + P(Q_i) > 1$}{
    \Comment*[l]{get minimum $P(C_i \cap Q_i)$}
    $P(C_i \cap Q_i) \leftarrow P(C_i) + P(Q_i) - P(C_i \cup Q_i)$ \;
    \Comment*[l]{max $P(C_i \cup Q_i)$ can be $1$}
    \Comment*[l]{$P(C_i \cap Q_i) \geq P(C_i) + P(Q_i) - 1$}
    \Comment*[l]{min $P(C_i \cup Q_i)$}
    $P(C_i \cap Q_i) \leftarrow P(C_i) + P(Q_i) - 1$ \;
    \If{$P(C_i \cap Q_i) > T_f$}{
        \Comment*[l]{fail guard-railing}
    }
    \Else{
        \Comment*[l]{pass guard-railing}
    }
}
\Else{
    \Comment*[l]{skip guard-railing}
}

\end{algorithm}

When a guard-rail condition assertion fails, the associated hot-fix is updated by operators to make the guard-rail criteria is met. It should be noted here that adding and updating hot-fixes is only a temporary solution because it takes away traffic from the RL policy and redirects it towards make-shift hand-crafted rules which hampers the scalability of the larger system. It is therefore crucial to start the process of properly assimilating the traffic handled by these rules back to the RL policy after the short-term mitigation.

\subsection{Remediation}
\label{sec:remediation}
As a part of a regular training cycle for off-policy learning, we optimize a loss function $L_0$. For simplicity of explanation, in this paper, we use the inverse propensity scoring (IPS) objective as an example for the case of contextual bandit formulation \citep{dudik2014doubly}:
\begin{equation}
    L_0 = \mathbb{E}_{X,a,r \sim \mathbb{D}} = -r  \frac{\Pi_\theta(a|X)}{\Pi_0(a|X)}.
    \label{eq:loss}
\end{equation}

\begin{figure}[t]
    \centering
        \includegraphics[width=0.9\linewidth]{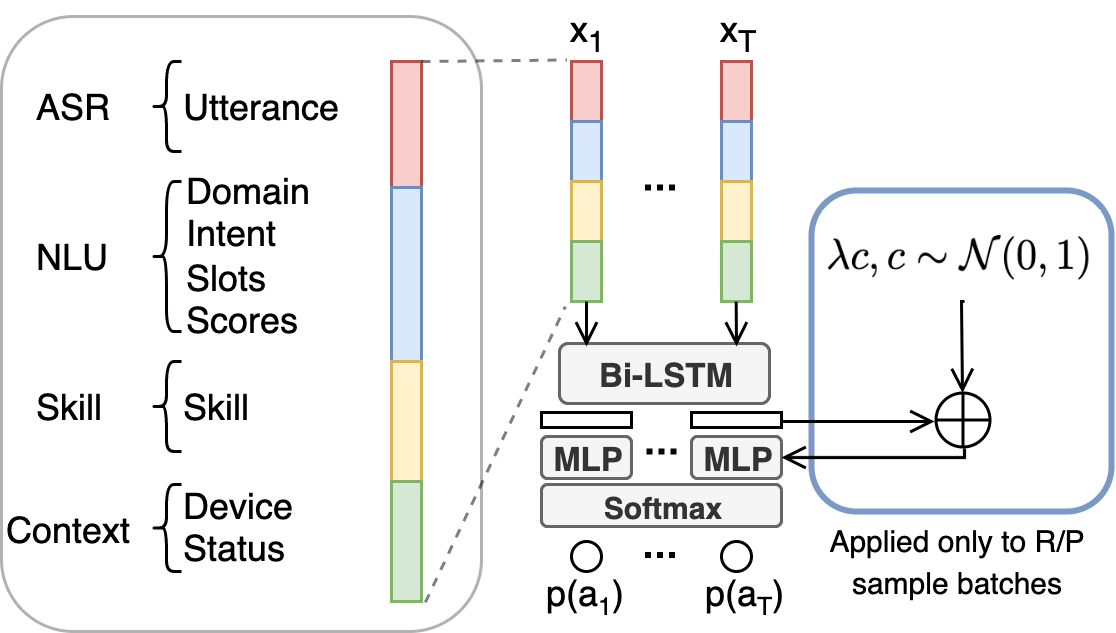}
        \caption{Model architecture used for the RL policy; augmented R/P sample batches are injected with gaussian noise during the forward pass at their hidden-layer representations as shown in the blue box.}
        \label{fig:model_arch}
\end{figure}

We inject R/P samples in the training loop 
to the regular training batches and replay them during each iteration. To improve the generalization and data efficiency of using the limited R/P data, we perform representation space data augmentation. This is done on a mini-batch of R/P samples using Gaussian noise injection during the forward pass on each hypothesis at hidden-layer representations as depicted in figure \ref{fig:model_arch}. It is further defined in the equation below where $\bar{\mathbf{x}}$ is the hidden space feature vector for hypothesis $\mathbf{x}$, $\bar{\mathbf{x}'}$ is the augmented sample vector, $j$ is the feature index and $\lambda$ is the noise scaling factor.

\begin{equation}
      \bar{\mathbf{x}'}_j = \bar{\mathbf{x}'}_j + \lambda c , c \sim \mathcal{N}(0, 1)
    \label{eq:noise}
\end{equation}

The auxiliary loss ($L_{RP}$) is computed from the regular loss objective ($L_0$) albeit on augmented data sampled from R/P dataset, $\mathbb{D}_{RP}$, represented as $\mathbb{D}_{RP}'$. When introducing the R/P samples as a part of the training data, we make adjustments such that the added samples discourage action replication for regression cases and encourage replication logged of actions for progression cases. To implement this, we reshape reward values such that regression and progression cases get the lowest and highest possible reward. We represent this reshaped reward via $r'$, and the auxiliary loss in equation \ref{eq:loss_aug}.
\begin{equation}
    L_{RP} = \mathbb{E}_{X,a,r' \sim \mathbb{D}_{RP}'} = -r'  \frac{\Pi_\theta(a|X)}{\Pi_0(a|X)}.
    \label{eq:loss_aug}
\end{equation}

Finally, we perform a weighted average of the auxiliary loss ($L_{RP}$) with the regular loss ($L_0$) using a weight term $\eta$ to get the overall loss as depicted in equation \ref{eq:final_loss}.
\begin{equation}
    L = (1-\eta)L_0  + (\eta)L_{RP} ,\;\; 0 < \eta < 1.
    \label{eq:final_loss}
\end{equation}

Additionally, we have parameters, $\alpha$ and $\beta$, that control the number of R/P samples per batch and number of augmentations to perform per R/P sample in the training loop respectively. Refer algorithm \ref{alg:aer} for more step by step details.

\begin{algorithm}[t]
\small
\caption{Augmented Exp. Replay}
\label{alg:aer}

\DontPrintSemicolon
\SetKwInOut{Input}{input}
\SetKwInOut{Output}{output}

\Input{$\mathbb{D}$ (dataset of logged interactions from $\Pi_0$), $\mathbb{D}_{RP}$ (dataset of R/P samples),\\ $\eta$ (train replay loss mix ratio), \\ $\alpha$ (\# R/P sample per regular batch), \\$\beta$ (\#  augmentations per R/P sample), \\$\lambda$ (noise scaling factor)}

$\mathbb{D} \leftarrow preprocess(\mathbb{D})$ \;

$\mathbb{D}_{RP} \leftarrow preprocess(\mathbb{D}_{RP})$ \;
$\mathbb{D}_{RP}' \leftarrow reshapeReward(\mathbb{D}_{RP})$ \;

\For{$d$ in $nextBatch(\mathbb{D})$}{
    \Comment*[l]{sample R/P batch with replacement}
    $d_{rp} = sampleBatch(\mathbb{D}_{RP}', size=\alpha * \beta)$ \;
    \Comment*[l]{loss on regular data batch}
    $L_0 \leftarrow loss(\Pi_\theta, d)$ \;
    \Comment*[l]{loss on rp data batch}
    $L_{RP} \leftarrow loss(\Pi_\theta, d_{rp}, noise=\lambda)$ \;
    \Comment*[l]{combine regular and R/P loss}
    $L \leftarrow (1-\eta) L_0 + (\eta)L' $ \;
    \Comment*[l]{use any optimizer $f$ for $\Pi_{\theta}$}
    $\theta \leftarrow f(\theta,\nabla_{\theta}L)$ \;
}

\end{algorithm}

\section{Experiments}
\subsection{Setup}
To evaluate the proposed remediation approach, we conducted online and offline experiments in real-world production settings. In this section, we use the term \textit{baseline policy} to refer to the approach suggested by \citet{kachuee2022scalable}. The proposed framework extend the baseline approach and henceforth referred as \textit{R/P policy}.

To simplify the comparisons, we follow the same model architecture and design choices as suggested by \citet{kachuee2022scalable}. In summary,
input to the model is a set of routing candidates, i.e., a combination of embedded ASR, NLU, and context vectors as well as skill embeddings. The output is the softmax-normalized propensity of selecting each candidate to handle the user request. The final model has about 12M trainable parameters consisting of a language model to encode utterance, embeddings for contextual signals, and fully-connected layers. 

To train and evaluate our models, we use logged data from a current production policy. The observed reward is based on a curated function of user satisfaction metrics. Our dataset consists of about 90M samples roughly divided into 75\% training, 12.5\% validation, and 12.5\% test hold-out sets covering tens of domains with imbalanced number of samples. Our R/P dataset consists of $\sim$50 samples and split into 67\% training and 33\% test hold-out sets containing roughly an equal number of regression and progression samples (collected over 10-15 reported defects). We ensure that each incident finds similar representation in both the train and test hold-out set. Data used in this work was de-identified to comply with our customer privacy guidelines. Also, due to confidentiality concerns, we are not able to share specifics about the historical regression incidents.



\subsection{Metrics\footnote{To comply with our privacy and business guidelines, in all instances, we only report relative and normalized results which do not represent the actual scales or metric values.}}
\subsubsection{Remediation Metrics}
We use \textit{remediation percentage} as a key metric to quantify the percentage of R/P samples with status FAIL that were directed back to the RL policy with status PASS in a single model update using the remediation approach shared in section \ref{sec:remediation}. In an ideal scenario we would expect this metric to be as high as possible. It is defined more concretely in equation~\ref{eq:reonboard} below where $C$ and $C'$ represent the sample statuses obtained from baseline and R/P policy respectively. 
\begin{equation}
    \frac{\sum\limits_{i=0}^{|\mathbb{D_{RP}|}} 1_{(C_i = FAIL)} - \sum\limits_{i=0}^{|\mathbb{D_{RP}|}} 1_{(C'_{i} = FAIL)}}{\sum\limits_{i=0}^{|\mathbb{D_{RP}|}} 1_{(C_{i} = FAIL)}}*100
    \label{eq:reonboard}
\end{equation}
\subsubsection{Deviation Metrics}
To validate that the auxiliary R/P loss is not having an adverse effect on other data segments, we track the deviation in \textit{decision replication rate} and the \textit{expected reward} for the remainder of traffic. In an ideal scenario we would expect both deviation metrics to be as small as possible. 

\subsection{Hyperparameters}
For the train replay loss mix ratio $\eta$ we use values from $\{0.02, 0.2\}$ and for noise variance $\lambda$ we use values from $\{0, 0.05, 1.0, 2.0, 3.0\}$ to find the best parameters based on the remediation percentage. We particularly note during an ablation that having no noise leads to poor generalization on the R/P hold out set. Consequently, we use a grid search for finding the best setting for the number of R/P samples per batch $\alpha \in \{2, 5, 10\}$ and number of augmentation per R/P sample $\beta \in \{1, 20, 50\}$ to find the best settings for each benchmark. Based on this search, we finally used $\eta$ as 0.2, $\alpha$ as 5, $\beta$ as 20 and $\lambda$ as 2.0. 

\subsection{Training Details}
For the baseline policy we trained each model for 8 epochs and take the best performing model based on the macro-averaged violation rate of added domain based constraints measured on the validation set. We used a cluster of 32 NVIDIA V100 GPUs to process a mini-batch size of 32K samples (1000 samples on each GPU). Each individual run took between 14 to 16 hours. During R/P policy training we added an augmented batch of 100 R/P samples ($\alpha = 5, \beta = 20$) to each GPU creating a further addition of 3200 samples to each mini-batch. Each experiment was run four times using different random seeds for weight initialization to report the mean and $\pm2$ standard deviation of each result.

\section{Results}
We conducted offline experiments and measured off-policy estimated impact of the proposed method on replication and reward metrics. For the estimating the expected reward, we used an IPS estimator.
On our training set we observed an average remediation percentage of 70.0\% (71.42\% for regression and 66.6\% for progression samples) indicating that the proposed approach leads to a high assimilation of the defective traffic back to RL policy. The number can also be interpreted as the normalized percentage of reduction in RP samples that used to be handled by the hot fixes and instead be handled correctly by the RL policy. Using this approach we were successfully able to absorb the entire hold out set to the RL policy and identify the potential to retire $\sim$70\% of the representative hot-fixes. 


Table~\ref{tab:offline_replication} shows the deviation percentage in decision replication rate and the off-policy estimated reward on the hold out dataset. We see negligible difference between both the policies indicating that the remediation has minimal side-effect on the remaining traffic segments.

\begin{table}[h]
\centering
\resizebox{\linewidth}{!}{
\renewcommand{\arraystretch}{1.0}
\begin{tabular}{l|cc}
\toprule
 \textbf{Offline}&  Replication (\%) & Expected Reward (\%) \\
 \hline
Baseline Policy & 98.31{\footnotesize$\pm${0.0005}} & 89.55{\footnotesize$\pm${0.0005}}\\
RP Policy & 98.31{\footnotesize$\pm${0.0071}} & 89.56{\footnotesize$\pm${0.0052}} \\
\hline
Deviation (\%) & 0.00{\footnotesize$\pm${0.0072}} & 0.01{\footnotesize$\pm${0.0054}} \\
\hline
\end{tabular}
}
\caption{Comparison of the overall replication and expected reward on our offline test set reported for the baseline and RP policies. 
}
\label{tab:offline_replication}
\end{table}


We then compared our proposed approach to the baseline on live production traffic in an online A/B based setup consisting of a large number of actual customers. The results in Table~\ref{tab:ab_results} show that, similar to our offline analysis, we observed minimal and non-statistically significant deviation in the measured reward between control and treatment. This further validates our claim that the proposed remediation has negligible impact on the remaining traffic segments. 

\begin{table}[h]
\centering
\resizebox{0.8\linewidth}{!}{
\renewcommand{\arraystretch}{1.0}
\begin{tabular}{l|c}
\toprule
 \textbf{Online}&  Measured Reward (\%) \\
 \hline
Baseline Policy &  87.81 \\
R/P Policy & 87.80 \\
\hline
Deviation (\%) & -0.01 (p-value 0.4)\\
\hline
\end{tabular}
}
\caption{Overall deviation between the baseline and the RP policy on the actual reward received during an online A/B. Here, p-value of 0.4 indicates no significant side-effect as a result of our proposed remediation.}
\label{tab:ab_results}
\vspace{-0.1in}
\end{table}

\section{Conclusion}
In this paper, we presented a method to leverage historical regressions reported by customers of a conversational AI to guard-rail against future recurrences of similar issues and to improve the trained policies to learn from such high-value experiences. In summary, the introduced method consists of curating a regression/progression dataset from historical incidences, logic to evaluate future polices on such data prior to the potential online deployment, performing guard-railing against deploying policies that pose a high risk of incident recurrences, and finally leveraging such a high-value dataset as a source of supervision during the training process to enable long-term behavior corrections. We conducted extensive online and offline experiments and deployed this work in a real-world production system to ensure serving best experience for our customers.

\section*{Limitations}
We believe a potential limitation of this work is its reliance of curated samples from historical incidents. Due to the complexity of real-world conversational agents, the decision to introduce a new sample to the R/P set requires human expert involvement which could be costly and pose challenges in terms of reliability. Another challenge we faced after the deployment of this framework was managing the life-cycle of the collected R/P samples. In a dynamic environment, a regression or progression pattern may lose relevance over time. Therefore, we find it challenging to re-actively deal with retirement of such historical samples.

\section*{Ethics Statement}
This work is centered on ensuring the best experiences are served by a conversational AI through learning and validation of customer initialed reports. Therefore, we do not assess any particular ethical risks associated with this work. However, one penitential though unlikely risk area would be human expert decisions for data collection to be biased on certain use-cases or interactions. We did not observe manifestation of such risk impacting our experiments and after the production deployment. Regarding human data handling practices, we ensured anonymity of data samples used in this study and did not reveal any specifics that would violate our internal policies or our customer privacy policies.

\bibliography{refs}

\begin{thebibliography}{14}
\expandafter\ifx\csname natexlab\endcsname\relax\def\natexlab#1{#1}\fi

\bibitem[{Balakrishnan et~al.(2018)Balakrishnan, Bouneffouf, Mattei, and
  Rossi}]{balakrishnan2018using}
Avinash Balakrishnan, Djallel Bouneffouf, Nicholas Mattei, and Francesca Rossi.
  2018.
\newblock Using contextual bandits with behavioral constraints for constrained
  online movie recommendation.
\newblock In \emph{IJCAI}, pages 5802--5804.

\bibitem[{Dud{\i}k et~al.(2014)Dud{\i}k, Erhan, Langford, and
  Li}]{dudik2014doubly}
Miroslav Dud{\i}k, Dumitru Erhan, John Langford, and Lihong Li. 2014.
\newblock Doubly robust policy evaluation and optimization.
\newblock \emph{Statistical Science}, 29(4):485--511.

\bibitem[{Hoffman et~al.(2014)Hoffman, Shahriari, and
  Freitas}]{hoffman2014correlation}
Matthew Hoffman, Bobak Shahriari, and Nando Freitas. 2014.
\newblock On correlation and budget constraints in model-based bandit
  optimization with application to automatic machine learning.
\newblock In \emph{Artificial Intelligence and Statistics}, pages 365--374.
  PMLR.

\bibitem[{Joachims et~al.(2018)Joachims, Swaminathan, and
  de~Rijke}]{joachims2018deep}
Thorsten Joachims, Adith Swaminathan, and Maarten de~Rijke. 2018.
\newblock Deep learning with logged bandit feedback.
\newblock In \emph{International Conference on Learning Representations}.

\bibitem[{Kachuee and Lee(2022)}]{kachuee2022constrained}
Mohammad Kachuee and Sungjin Lee. 2022.
\newblock Constrained policy optimization for controlled self-learning in
  conversational ai systems.
\newblock \emph{arXiv preprint arXiv:2209.08429}.

\bibitem[{Kachuee et~al.(2022)Kachuee, Nam, Ahuja, Won, and
  Lee}]{kachuee2022scalable}
Mohammad Kachuee, Jinseok Nam, Sarthak Ahuja, Jin-Myung Won, and Sungjin Lee.
  2022.
\newblock Scalable and robust self-learning for skill routing in large-scale
  conversational ai systems.
\newblock \emph{Annual Conference of the North American Chapter of the
  Association for Computational Linguistics (NAACL)}.

\bibitem[{Kachuee et~al.(2021)Kachuee, Yuan, Kim, and Lee}]{kachuee2021self}
Mohammad Kachuee, Hao Yuan, Young-Bum Kim, and Sungjin Lee. 2021.
\newblock Self-supervised contrastive learning for efficient user satisfaction
  prediction in conversational agents.
\newblock In \emph{Proceedings of the 2021 Conference of the North American
  Chapter of the Association for Computational Linguistics: Human Language
  Technologies}, pages 4053--4064.

\bibitem[{Karampatziakis et~al.(2019)Karampatziakis, Kochman, Huang, Mineiro,
  Osborne, and Chen}]{karampatziakis2019lessons}
Nikos Karampatziakis, Sebastian Kochman, Jade Huang, Paul Mineiro, Kathy
  Osborne, and Weizhu Chen. 2019.
\newblock Lessons from contextual bandit learning in a customer support bot.
\newblock \emph{arXiv preprint arXiv:1905.02219}.

\bibitem[{Ke et~al.(2022)Ke, Kachuee, and Lee}]{ke2022domain}
Zixuan Ke, Mohammad Kachuee, and Sungjin Lee. 2022.
\newblock Domain-aware contrastive knowledge transfer for multi-domain
  imbalanced data.
\newblock In \emph{Proceedings of the 12th Workshop on Computational Approaches
  to Subjectivity, Sentiment \& Social Media Analysis}, pages 25--36.

\bibitem[{Li et~al.(2021)Li, Park, Dara, Nam, Lee, Kim, Matsoukas, and
  Sarikaya}]{li2021neural}
Han Li, Sunghyun Park, Aswarth Dara, Jinseok Nam, Sungjin Lee, Young-Bum Kim,
  Spyros Matsoukas, and Ruhi Sarikaya. 2021.
\newblock Neural model robustness for skill routing in large-scale
  conversational ai systems: A design choice exploration.
\newblock \emph{arXiv preprint arXiv:2103.03373}.

\bibitem[{Lopez et~al.(2021)Lopez, Dhillon, and Jordan}]{lopez2021learning}
Romain Lopez, Inderjit~S Dhillon, and Michael~I Jordan. 2021.
\newblock Learning from extreme bandit feedback.
\newblock \emph{Proc. Association for the Advancement of Artificial
  Intelligence}.

\bibitem[{Park et~al.(2020)Park, Li, Patel, Mudgal, Lee, Kim, Matsoukas, and
  Sarikaya}]{park2020scalable}
Sunghyun Park, Han Li, Ameen Patel, Sidharth Mudgal, Sungjin Lee, Young-Bum
  Kim, Spyros Matsoukas, and Ruhi Sarikaya. 2020.
\newblock A scalable framework for learning from implicit user feedback to
  improve natural language understanding in large-scale conversational ai
  systems.
\newblock \emph{arXiv preprint arXiv:2010.12251}.

\bibitem[{Sarikaya(2017)}]{sarikaya2017technology}
Ruhi Sarikaya. 2017.
\newblock The technology behind personal digital assistants: An overview of the
  system architecture and key components.
\newblock \emph{IEEE Signal Processing Magazine}, 34(1):67--81.

\bibitem[{Swaminathan et~al.(2016)Swaminathan, Krishnamurthy, Agarwal,
  Dud{\'\i}k, Langford, Jose, and Zitouni}]{swaminathan2016off}
Adith Swaminathan, Akshay Krishnamurthy, Alekh Agarwal, Miroslav Dud{\'\i}k,
  John Langford, Damien Jose, and Imed Zitouni. 2016.
\newblock Off-policy evaluation for slate recommendation.
\newblock \emph{arXiv preprint arXiv:1605.04812}.

\end{thebibliography}
\bibliographystyle{acl_natbib}

\end{document}